\documentclass[runningheads, captions=tableheading]{llncs}

\usepackage[preprint,year=2024,ID=7502]{eccv}

\usepackage{eccvabbrv}

\usepackage{graphicx}
\usepackage{booktabs}

\usepackage[accsupp]{axessibility}  

\usepackage{booktabs}      
\usepackage{multirow}      
\usepackage{pifont}      
\usepackage{float}      
\newcommand{\xmark}{\ding{55}}
\usepackage[]{algorithm2e}
\usepackage{color, colortbl}

\definecolor{FadedGray}{gray}{0.95}
\definecolor{LightGray}{gray}{0.9}
\definecolor{MediumGray}{gray}{0.775}
\definecolor{HeavyGray}{gray}{0.65}

\usepackage[pagebackref,breaklinks,colorlinks,citecolor=eccvblue]{hyperref}

\usepackage{orcidlink}

\begin{document}

\title{Mind the map! Accounting for existing maps when estimating online HDMaps from sensors.} 

\titlerunning{Mind the map!}

\author{Rémy Sun\inst{1} \and
Li Yang\inst{1} \and
Diane Lingrand\inst{1} \and
Frédéric Precioso\inst{1}}

\authorrunning{R. Sun et al.}

\institute{Université Côte d’Azur, Inria, CNRS, I3S, Maasai, Nice, France \\
    \{firstname.lastname\}@univ-cotedazur.fr
}

\maketitle
\begin{abstract}
  While HDMaps are a crucial component of autonomous driving, they are expensive to acquire and maintain. Estimating these maps from sensors therefore promises to significantly lighten costs. These estimations however overlook existing HDMaps,
with current methods at most geolocalizing low quality maps or considering a
general database of known maps. In this paper, we propose to account for
existing maps of the precise situation studied when estimating HDMaps. We
identify 3 reasonable types of useful existing maps (minimalist, noisy, and
outdated). We also introduce MapEX, a novel online HDMap estimation framework
that accounts for existing maps. MapEX achieves this by encoding map elements
into query tokens and by refining the matching algorithm used to train classic
query based map estimation models. We demonstrate that MapEX brings significant
improvements on the nuScenes dataset. For instance, MapEX - given noisy maps -
improves by 38\% over the MapTRv2 detector it is based on and by 8\% over the
current SOTA.
\end{abstract}
\section{Introduction}
\label{sec:intro}

Autonomous Driving \cite{kuutti2020survey, Gao_2020_CVPR} represents a complex
problem that promises to significantly change how we interact with
transportation. While full vehicle automation still seems quite a ways away
\cite{teng2020TIV}, partially autonomous vehicles now populate a number of road
systems in the world \cite{kuutti2020survey}. These vehicles need to process a
wealth of information to function, from the raw sensor data \cite{Gu_2023_CVPR}
to elaborate maps of road networks \cite{Gao_2020_CVPR, liang2020learning}.

\begin{figure}
  \begin{subfigure}{0.075\linewidth}
    \centering
    \includegraphics[width=\linewidth]{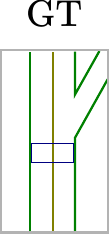}
  \end{subfigure}
  \hfill
  \begin{subfigure}{0.425\linewidth}
    \centering
    \includegraphics[width=\linewidth]{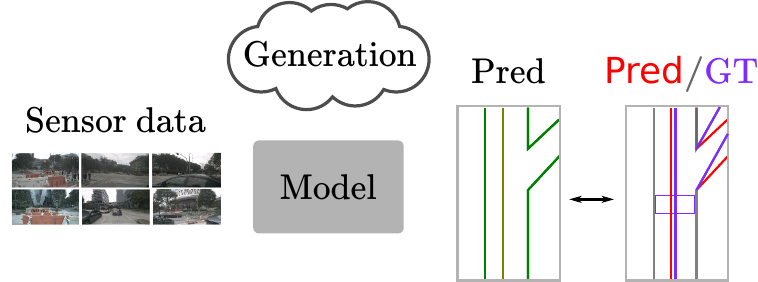}
    \caption{With only sensor data.}
    \label{fig:intro_pw}
  \end{subfigure}
  \hfill
  \begin{subfigure}{0.425\linewidth}
    \centering
    \includegraphics[width=\linewidth]{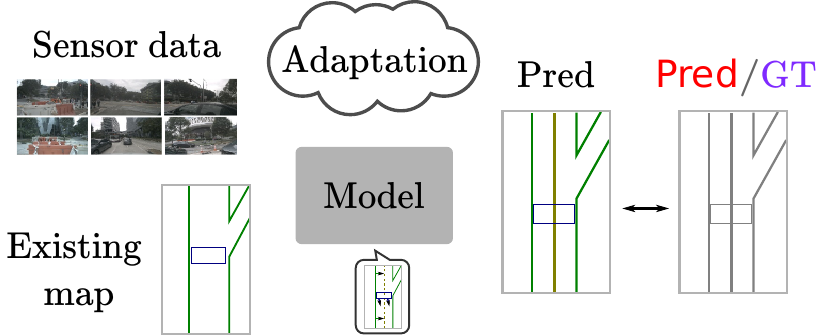}
    \caption{With map information (ours).}
    \label{fig:intro_mapex}
  \end{subfigure}
  \caption{We propose to use existing map information - even if inexact - to
    estimate better online HDMaps from sensor inputs. In doing so, we simplify
    the problem from generating maps using only sensors to re-using
    available existing maps aided by sensors.}
  \label{fig:intro}
\end{figure}

High Definition maps (HDMaps), in particular, represent a crucial component of
the research on self-driving cars \cite{Gao_2020_CVPR, elghazaly2023ITS} (see
\cref{fig:intro} for a few simple examples of maps, with road boundaries
represented by green polylines, lane dividers by lime polylines and pedestrian
crossings by blue polygons). Although maps are not a typical input of neural
networks, they contain necessary information to help the car understand the
world it must navigate. As such, significant efforts have gone into
incorporating this new type of data into solutions \cite{Gao_2020_CVPR,
park2023leveraging}. These efforts have shown HDMaps' remarkable
benefits both for fundamental problems like Object Detection
\cite{hausler2023displacing} to precise trajectory forecasting problems
\cite{liang2020learning, park2023leveraging}.

These maps are however expensive to acquire and maintain, requiring precise data
acquisition and exacting human labeling \cite{elghazaly2023ITS,
jeong2022tutorial}. There has therefore been a strong push
\cite{can2021structured, liao2023maptr} to approximate HDMaps from sensor
data. Seminal work \cite{liu2022vectormapnet} has proved even
rough estimated maps useful for trajectory forecasting.

While recent methods like MapTRv2 \cite{liao2023maptrv2} have become proficient
at generating online HDMaps from raw sensors, we feel they overlook very useful
and nearly always available data: existing maps. We posit here that outdated or lower quality maps should usually be available and prove to significantly improve
the estimated HDMaps as illustrated in \cref{fig:intro}. Indeed, even
``map-free'' models tend to use lower-quality satnav maps \cite{wayve}, and
estimated maps could always be available as long as a vehicle went through a
place once.

In this paper, we explore the central postulate that \textbf{even inaccurate existing maps improve the estimation of HDMaps} from raw sensors.
After providing some context on our method and the field
in \cref{sec:rw}, we propose two distinct technical contributions: In \cref{sec:scenarios},
we outline reasonable scenarios under which an inaccurate map can be available
along with practical implementations, and in \cref{sec:mapex}, we propose MapEX, an architecture that can generate HDMaps from sensor
data while accounting for existing map information. Finally, we present results in \cref{sec:exp} with experiments on the nuScenes dataset
\cite{nuscenes2019}.\\

\noindent \textit{Contributions} We detail the following three contributions in this paper:
\begin{itemize}
\item We propose to \textbf{account for existing map information} when estimating
online HDMaps from sensor data.
\item We discuss \textbf{reasonable scenarios} under which existing maps are not perfect. We also provide realistic implementations of these scenarios and the code for the nuScenes dataset.
\item We introduce \textbf{MapEX}, a new query based HDMap estimation framework that can \textbf{incorporate existing map information when estimating an online HDMap from sensors}. In
particular, we introduce with MapEX both a novel way to incorporate existing map
information with non-learnable existing (EX) queries, and a way to ensure the model uses this
information by pre-attributing predictions to known ground truth correspondences during training.
\end{itemize}

\noindent \textit{Impact} We believe these contributions are of
substantial interest to the field as:
\begin{itemize}
\item MapEX drastically \textbf{lowers the bar} needed to obtain good and cost-effective map
estimations. In one scenario where we use HDMaps with noisy (or ``shifted'') map
element positions, for instance, MapEX reaches a 84.8\% mAP score which is an
improvement of 38\% over the MapTRv2 detector it is based on. This is also a
8\% improvement over the state-of-the-art set by MapNeXt \cite{li2024mapnext}
using a
foundation model image backbone \cite{wang2023internimage} (vs. our ResNet-50 backbone). As
estimated maps become more complex, MapEX will become more and more crucial to
good and cost-effective performance.
\item MapEX is the \textbf{first method to directly integrate an existing map} corresponding precisely to the sensor data from the given location (i.e. the input sample) to guide online
  HDMap estimation. To the best of our knowledge, this is a blind spot of the literature, with previous works only considering geolocalized satellite
  maps \cite{gao2023complementing}, SDMaps \cite{anonymous2023pmapnet, luo2023augmenting}, or trying to retrieve an existing map
  similar to the sample from a pre-computed database \cite{wu2023pix2map}.
  None of these methods leverage sensor data to correct a flawed existing map.
\item We provide the implementation of existing map scenarios for online
  HDMap estimation following three realistic scenarios. Our implementations
  very importantly provide both (flawed) existing HDMaps and the true HDMap for
  each sample. This is not the case in the existing \textit{Trust but Verify} \cite{Lambert21neurips} map change
  detection dataset which only provides the existing HDMap and a label as to
  whether a change has occurred.
\end{itemize}

\noindent \textit{Map nomenclature} We work on \textbf{local} $30m\times60m$ maps restricted to a sample's surroundings.
\textbf{True} maps are the ground truth maps, \textbf{existing} maps are the maps available as
inputs, \textbf{predicted} maps are the maps estimated by a model.


\section{Related Work}
\label{sec:rw}

We provide here some brief context on HDMaps in autonomous driving. We begin by
discussing HDMap's use in trajectory forecasting, before discussing their
acquisition. We then discuss online HDMap estimation itself.

\textit{HDMaps for trajectory forecasting }
Autonomous Driving requires a lot of information about the world
vehicles are to navigate. This information is typically embedded in rich HDMaps
given as input to modified neural networks \cite{gulzar2021survey,
park2023leveraging}. HDMaps have proven critical to the performance of a number
of modern methods in trajectory forecasting \cite{park2023leveraging,
deo2021multimodal} and other applications \cite{hausler2023displacing}. In
trajectory forecasting in particular, it is remarkable that some methods
\cite{liu2021multimodal, liu2023laformer} explicitly reason on a representation
of the HDMap and therefore absolutely require access to a HDMap \cite{Sun_2023_ICCV}. 
\cite{liu2022vectormapnet} reports a 10\% drop in performance for
a common forecasting technique \cite{liu2021multimodal} when applied without an
informative HDMap. \cite{xu2023motion} reports even more dramatic drops in
performance for other well known methods.

\textit{HDMap acquisition and maintenance }
Unfortunately, HDMaps are expensive to acquire and maintain \cite{elghazaly2023ITS, jeong2022tutorial}. While HDMaps used in forecasting are
only a simplified version containing map elements (lane dividers, road
boundaries, ...) \cite{li2022hdmapnet, liao2023maptrv2} and leave out much of the
complex information in full HDMaps \cite{elghazaly2023ITS}, they still require
exceedingly precise measurements (on the scale of tens of centimeters)
\cite{elghazaly2023ITS}. A number of companies have therefore been moving
towards a less exacting standard with Medium Definition Maps (MDMaps) \cite{mdmaps},
or even simpler Standard Definition Maps (SDMaps) such as satellite navigation maps, Google Maps, etc \cite{wayve}.
Crucially, MDMaps - with their precision of a few meters - would be a good
example of an existing map giving valuable information for the online HDMap
generation process. Our map \textcolor{Mulberry}{\textbf{Scenario 2a}} explores an approximation of MDMaps.

\textit{Online HDMap estimation from sensors }
Online HDMap estimation \cite{can2021structured} has therefore emerged as a promising alternative to manually curated HDMaps. While some works \cite{can2021structured, xu2022centerlinedet, can2022topology} focus on predicting virtual map elements, i.e. lane centerlines, the standard formulation introduced by \cite{li2022hdmapnet} focuses on more
visually recognizable map elements: lane dividers, road boundaries and
pedestrian crossings. Probably because visual elements are easier to detect by
sensors, this latter formulation has seen rapid progress over the last years
\cite{liu2022vectormapnet, liao2023maptr, Ding_2023_ICCV}. Interestingly, the
latest such method - MapTRv2 \cite{liao2023maptrv2} - does offer an auxiliary setting for detecting virtual lane centerlines. This suggests a natural convergence
towards the more complex settings comprising a multitude of additional map
elements (traffic lights, ...) \cite{li2023topology, wang2023openlanev2}.
Nevertheless, \textbf{the standard formulation from \cite{li2022hdmapnet} remains the gold standard} when evaluating the usefulness of additional information such as 
learned global feature maps \cite{xiong2023neural}, satellite views
\cite{gao2023complementing}, or SDMaps \cite{anonymous2023pmapnet}. We thus keep to this standard problem formulation to demonstrate the use of existing map information.

Our work is adjacent to the commonly studied map change detection problems
\cite{pannen2019hd, bu2023towards} that aim to detect a change in a map (e.g.
crossings). While rooted in more classical statistical techniques
\cite{pannen2019hd}, a few efforts have been made to adapt them to deep learning
\cite{heo2020hd, bu2023towards}. Notably, the Argoverse 2 \textit{Trust but Verify} (TbV)
dataset \cite{Lambert21neurips} was recently proposed for this problem (see
Appendix \cref{app:tbv}). This however differs substantially from our
approach as we do not try to correct small mistakes on an existing map after
aggregating from a fleet of vehicles \cite{pannen2020how, kim2021hd}. Instead we
aim to generate accurate online HDMaps with the help of an existing -
possibly very different - map, which is made possible by the modern
online HDMap estimation problem. Therefore, \textbf{we do not only correct small mistakes in maps but propose a more expressive framework that accommodates any change} (e.g. distorted lines, very noisy elements).

\section{What Kind of Existing Map Could We Use? }
\label{sec:scenarios}


We make the central claim that accounting for existing maps would benefit online HDMaps estimation. We argue here there are many reasonable scenarios under which imperfect existing maps
can appear. After defining our HDMap representations in
\cref{sec:mapdef} and our general approach in \cref{sec:mapmodex}, we
consider three main possibilities: only road boundaries are available
(\cref{sec:minimalist}), the maps are noisy (\cref{sec:noisy}), or they
have changed substantially (\cref{sec:outdated}).

\subsection{HDMap Representation For Online HDMap Estimation}
\label{sec:mapdef}

\begin{figure}[t]
\centering
\begin{subfigure}{0.24\linewidth}
  \includegraphics[width=.7\linewidth]{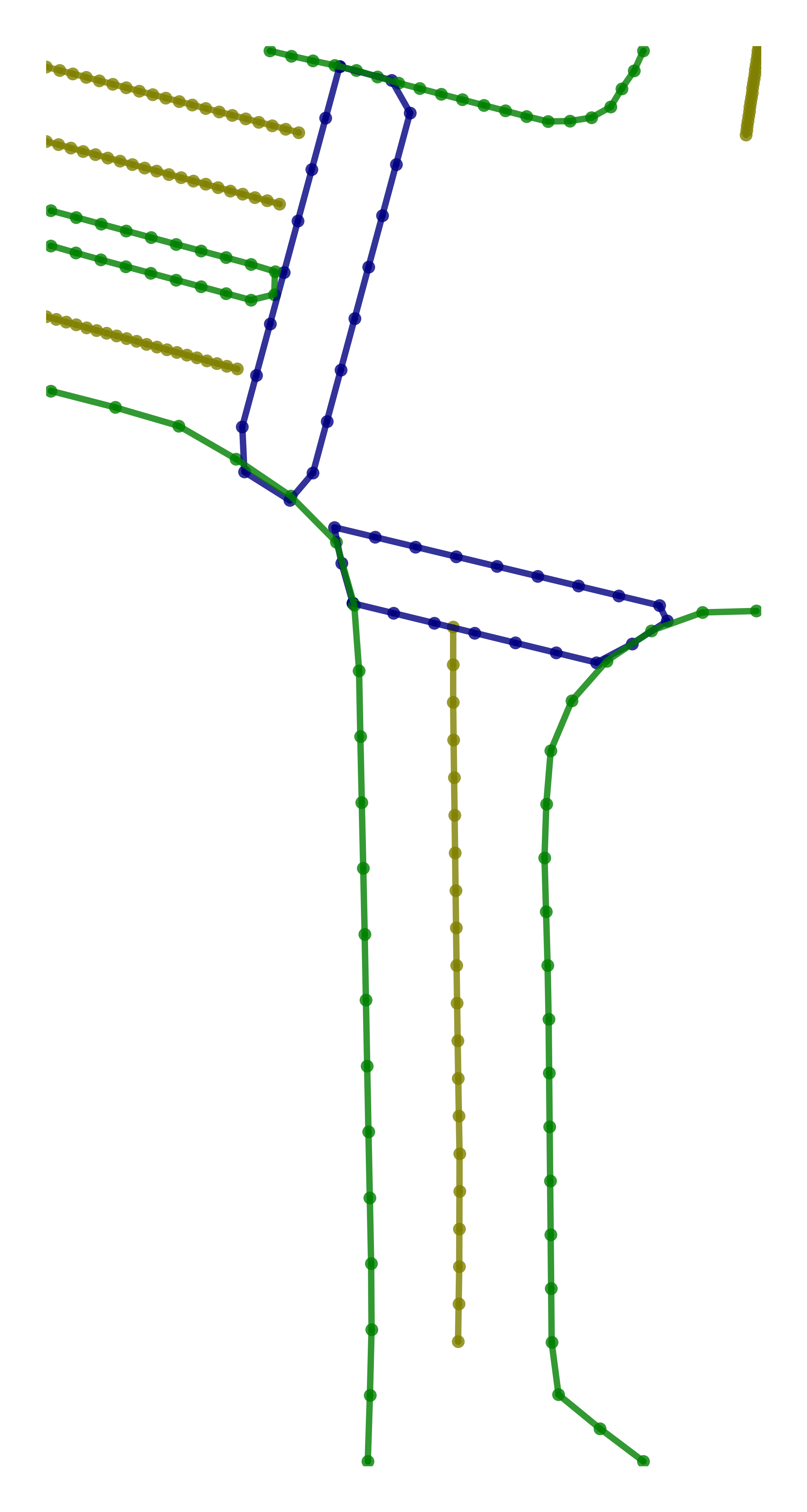}
  \caption{Ground Truth: \\True HDMap.}
  \label{fig:scenariogt}
\end{subfigure}
\hfill
\begin{subfigure}{0.24\linewidth}
  \includegraphics[width=.7\linewidth]{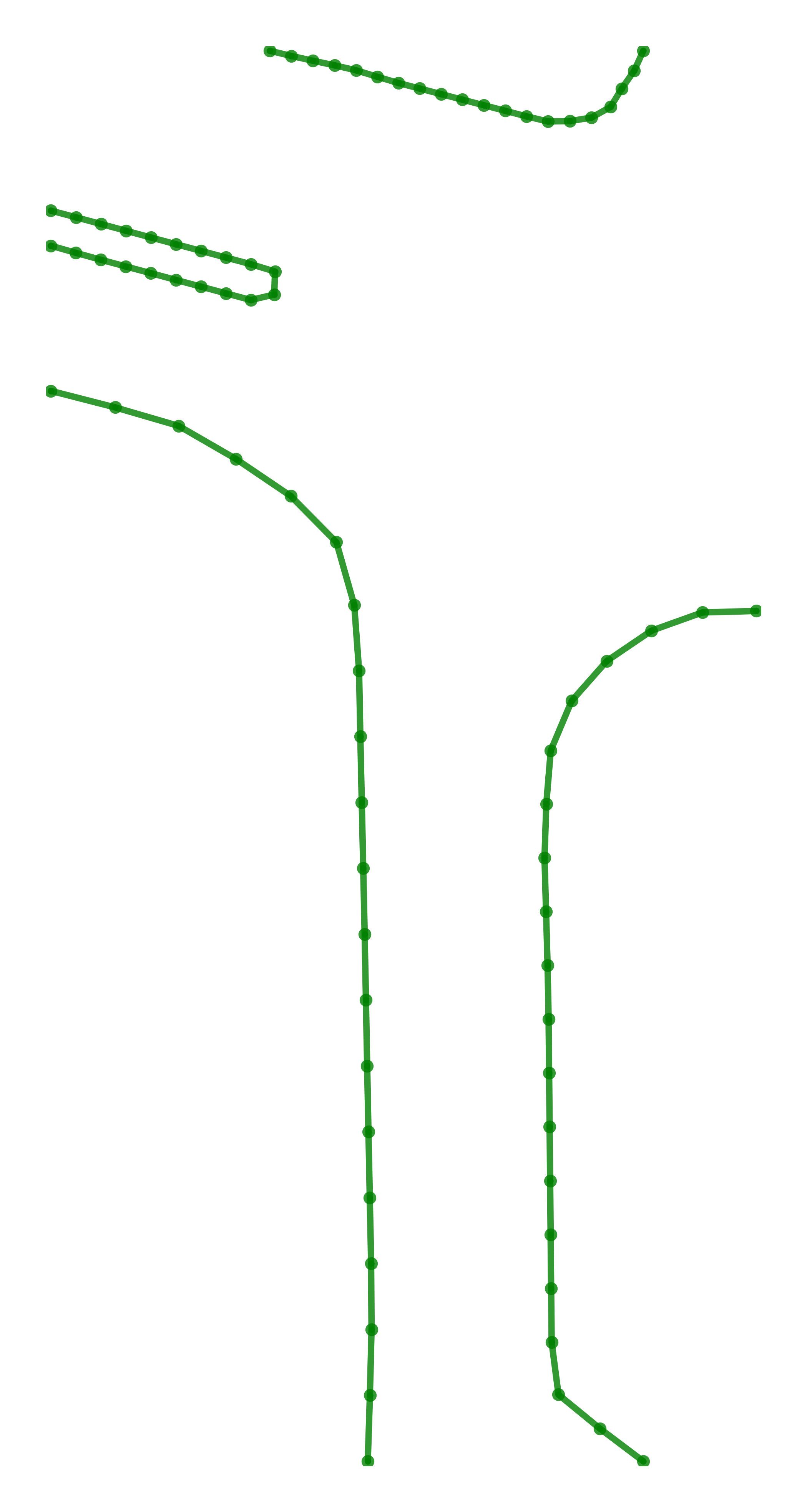}
  \caption{\textcolor{PineGreen}{\textbf{Scenario \hyperref[sec:minimalist]{1}}:} \\only road boundaries.}
  \label{fig:scenario1}
\end{subfigure}
\hfill
\begin{subfigure}{0.24\linewidth}
  \includegraphics[width=.7\linewidth]{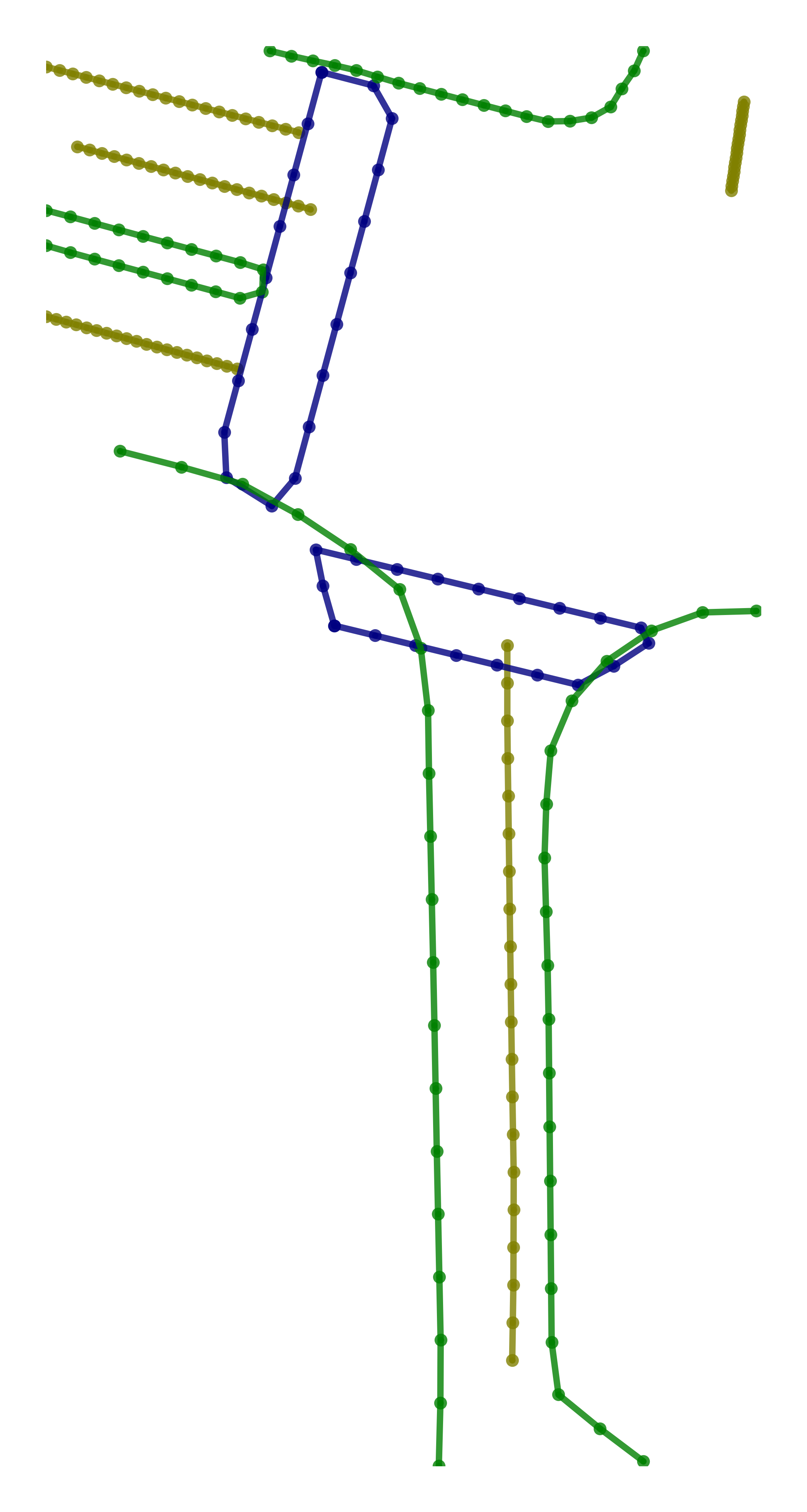}
  \caption{\textcolor{Mulberry}{\textbf{Scenario \hyperref[sec:noisy]{2a}}:} \\noisy element shift.}
  \label{fig:scenario2}
\end{subfigure}
\hfill
\begin{subfigure}{0.24\linewidth}
  \includegraphics[width=.7\linewidth]{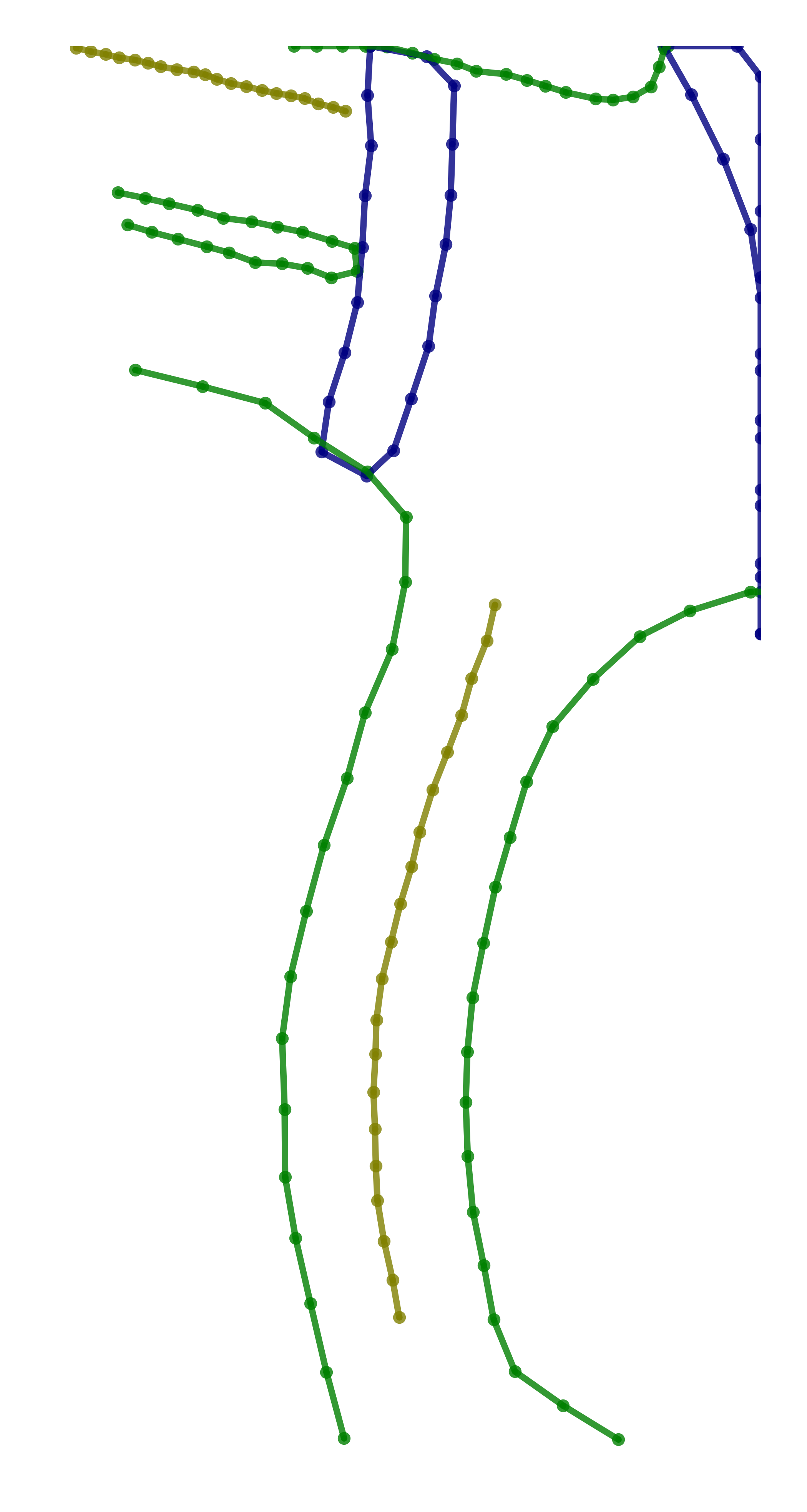}
  \caption{\textcolor{RoyalBlue}{\textbf{Scenario \hyperref[sec:outdated]{3a}}:}\\ outdated maps.}
  \label{fig:scenario3}
\end{subfigure}
\caption{Examples of HDMaps generated by MapModEX.}
\label{fig:scenarios}
\end{figure}

We adopt the standard format used for online HDMap estimation from
sensors \cite{li2022hdmapnet, liao2023maptr}. We consider HDMaps to be made of
three types of polylines (as represented on \cref{fig:scenariogt}):
road boundaries, lane dividers and pedestrian crosswalks with same colors as
previously green, lime, and blue respectively. We follow \cite{liao2023maptr} by
representing these polylines as sets of 20 evenly spaced points for our map
generator, with upsampled versions for evaluation.

While complete HDMaps are much more complex \cite{elghazaly2023ITS} and more
intricate representations have been proposed \cite{can2022topology}, the aim of
this work is to study how to account for existing map information. As such we
restrict ourselves to the most commonly studied formulation (road boundaries, lane dividers and pedestrian crosswalks), but our approach will be directly applicable to the prediction of more map elements
\cite{can2022topology}, finer polylines \cite{Ding_2023_ICCV, qiao2023end} or
rasterized objectives \cite{zhang2023online}.

\subsection{MapModEX: Simulating Imperfect Maps}
\label{sec:mapmodex}

As acquiring genuine imprecise maps for standard map acquisition datasets (e.g.
nuScenes) would be costly and time consuming, \textbf{we synthetically generate imprecise existing maps from true HDMaps}.

We develop MapModEX, a standalone map modification library. It takes nuScenes
map files and sample records, and for each sample outputs polyline coordinates
for dividers, boundaries and pedestrian crosswalks in a given patch, around the
ego vehicle. Importantly, our library provides the ability to modify these
polylines to reflect various modifications: removal of map elements, addition,
shifting of pedestrian crossings, noise addition to point coordinates, map shift, map rotation and
map warping. MapModEX will be made available after publication to facilitate
further research into incorporating existing maps into online HDMap acquisition
from sensors.

We implement \textbf{three challenging scenarios}, outlined next, using our MapModEX
package, generating for each sample 10 variants of scenarios
\textcolor{Mulberry}{\textbf{2a}}, \textcolor{Mahogany}{\textbf{2b}}, \textcolor{RoyalBlue}{\textbf{3a}}
and \textcolor{Blue}{\textbf{3b}} (scenario \textcolor{PineGreen}{\textbf{1}}
only admits one variant). We chose to work with a fixed set of modified maps to
reduce online computation costs during training and to reflect real situations
where only a finite number of map variants might be available.

\subsection{\textcolor{PineGreen}{\textbf{Scenario 1}}: Only Road Boundaries Are Available}
\label{sec:minimalist}

A first scenario is one where only a bare HDMap (with road boundaries but without divider or pedestrian
crossing) is available as shown on \cref{fig:scenario1}. Road boundaries
are more often associated with 3D physical landmarks (e.g. edge of sidewalk)
whereas dividers and pedestrian crossings are generally denoted by flat painted
markings that are easier to miss. Moreover, pedestrian crossings and lane
dividers are fairly commonly displaced by construction works or road deviations, or even partially hidden by tire tracks.

As such, it is reasonable to use HDMaps with \textbf{only road boundaries}.
This would have the benefit of reducing annotators costs by only asking
annotators to label road boundaries. Furthermore, less precise equipment and
less updates might be required to situate only road boundaries.

\textit{Implementation } \textcolor{PineGreen}{\textbf{Scenario 1}} is straightforward: we remove the
divider and pedestrian crossings from available HDMaps.

\subsection{\textcolor{Mulberry}{\textbf{Scenarios 2a}} and \textcolor{Mahogany}{\textbf{2b}}: Maps Are Noisy}
\label{sec:noisy}

A second plausible case involves noisy maps as shown on
\cref{fig:scenario2}. A weak point of existing HDMaps is the need for high
precision (in the order of a few centimeters), which puts a significant strain
on their acquisition and maintenance \cite{elghazaly2023ITS}. In fact, a key
difference between HDMaps and the emergent MDMaps standard lies in a lower
precision (a few centimeters vs. a few meters).

We therefore propose to work with \textbf{noisy HDMaps to simulate a cheaper acquisition process or a shift to the MDMaps standard}. More interestingly, acquiring HDMaps with automatic methods
could also lead to noisy maps. Although methods like MapTRv2 have
reached very impressive performance, they are not yet completely precise: the Mean Average Precision of predicted maps struggles
to reach even 70\%.

\textit{Implementation } We propose two possible implementations of these noisy HDMaps to reflect the various conditions under
which we might be lacking precision. In a first \textcolor{Mulberry}{\textbf{Scenario 2a}}, we propose
a \textbf{shift-noise setting} where we add noise from a Gaussian distribution with standard deviation of 1 meter on the localization of each map element. This has the effect
of applying a uniform translation to the points defining a given map element (divider, boundary, crosswalk). Such a setting should be a good approximation of situations where
human annotators provide quick imprecise annotations from noisy data. We chose a
standard deviation of 1 meter to reflect MDMaps standards of being precise up to
a few meters \cite{mdmaps}.

We then test our approach with a very challenging \textbf{pointwise-noise setting} in 
\textcolor{Mahogany}{\textbf{Scenario 2b}}: for each ground truth point - keeping in mind a map element is made up of 20 such points - we sample noise from a Gaussian
distribution with standard deviation of 5 meters and add it to the point
coordinates. This provides a worst case approximation of a possible situation -
in the future - where models automatically acquire maps or where
very imprecise localizations are used.

\subsection{\textcolor{RoyalBlue}{\textbf{Scenarios 3a}} and \textcolor{Blue}{\textbf{3b}}: Maps Have Substantially Changed}
\label{sec:outdated}

The final scenario we consider is one where we have access to \textbf{old maps that used
to be accurate} (see \cref{fig:scenario3}). As noted in
\cref{sec:minimalist}, it is fairly common for painted markers like
pedestrian crossings to be displaced from time to time. Furthermore, it is not
uncommon for cities to substantially remodel some problematic intersection or
renovate districts to accommodate traffic increase by a new attraction
\cite{plachetka2020terminology}.

It is therefore interesting to use existing HDMaps that are valid on their own but differ from the true HDMaps in significant ways. These maps should often appear when
the HDMaps are only updated by the maintainer every few years to cut down on
costs. In that case, the available maps would still provide some information on
the world but might not reflect temporary or recent changes.

\textit{Implementation } We approximate this situation by applying strong
changes to true HDMaps in our \textcolor{RoyalBlue}{\textbf{Scenario 3a}}. We delete
50\% of the pedestrian crossings and lane dividers, add a few
pedestrian crossings (half the amount of the remaining crossings) and finally
apply a small warping distortion to the map.

However, it is important to note that a substantial amount of the global map
will remain unchanged over time. We account for that in our
\textcolor{Blue}{\textbf{Scenario 3b}}, where we randomly choose (with
probability p=0.5) to keep the true HDMap instead of the perturbed existing HDMap.

\section{MapEX: Accounting for EXisting Maps}
\label{sec:mapex}

We propose MapEX (see \cref{fig:overview}), a novel framework for online
HDMap estimation. It follows the classic query based online HDMap estimation
framework \cite{liu2022vectormapnet, liao2023maptr} and processes existing map
information thanks to two key modules: a \textbf{map query encoding}
module (see \cref{sec:query}) and a \textbf{pre-attribution} of
predictions to known ground truth correspondences for training (see \cref{sec:matching}). We
also discuss an optional change detection module in Appendix \cref{app:change}. Since our implementation is built upon the
state-of-the-art MapTRv2 \cite{liao2023maptrv2}, it will translate to most
methods \cite{liu2022vectormapnet, Ding_2023_ICCV, liao2023maptr}

\subsection{Overview}
\label{sec:overview}

\paragraph{Base framework}

\begin{figure}[t]
  \centering
  \includegraphics[width=0.9\linewidth]{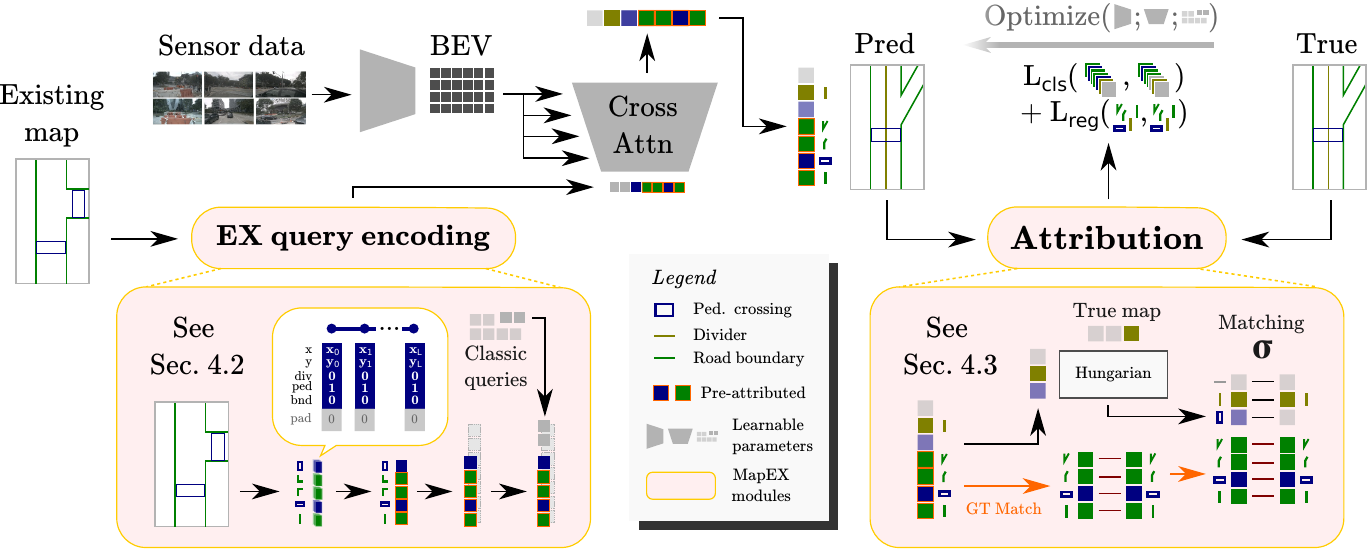}
  \caption{\textbf{Overview of our MapEX method} (see \cref{sec:mapex}). We
    add two modules (\textbf{EX query encoding}, \textbf{Attribution}) to the
    classic query based pipeline. Map elements
    are encoded into EX queries that are added to classic learned queries. These
  queries are then decoded using sensor data to yield map elements.
  Pre-attribution of known prediction to true map elements helps train stronger
models.}
  \label{fig:overview}
\end{figure}

The classic query based framework relies on a few trainable components
(gray elements on \cref{fig:overview}): a sensor to BEV
\textbf{encoder}, learnable detection \textbf{queries} and a map
\textbf{decoder}. It takes sensor inputs, processes them
and outputs predicted map elements (see Appendix \cref{app:classic}).

It starts by taking \textbf{sensor inputs} (cameras and/or LiDAR), and
\textbf{encodes} them into a Bird's Eye View (BEV) representation to serve as
sensor features. The map itself is obtained using a \textbf{DETR-like}
\cite{carion2020end} detection scheme to detect the map elements ($N$ at most).
It passes $N\times L$ \textbf{learned query} tokens ($N$ being
the maximum number of detected elements, $L$ the number of points predicted for
an element, with $L=20$ in this paper) into a transformer map \textbf{decoder} that feeds sensor
information to the query tokens using \textbf{cross-attention} with the BEV features. The
\textbf{decoded queries} are then translated into map element \textbf{coordinates} by linear layers along with a class prediction (including an extra
background class) such that groups of $L$ queries represent the $L$ points of a
map element. \textbf{Training} is done by finding a
\textbf{matching} $\sigma$ between predicted map elements
$\{\hat{y}_i=(\hat{c}_i,\hat{p}_i)\}_i$ and true (ground truth) map elements
$\{y_i=(c_i,p_i)\}_i$ (possibly padded with empty elements) using some variant
of the Hungarian algorithm \cite{kuhn1955hungarian, crouse2016on}. Once matched,
the model is optimized using a regression loss $\mathcal{L}_{reg}$ (for
coordinates) and classification (for element classes) losses
$\mathcal{L}_{cls}$:

\begin{equation}
  \mathcal{L}(\hat{y},y) = \frac{1}{N} \sum_{i=0}^{N-1} \mathcal{L}_{cls}(\hat{c}_{\sigma(i)}, c_{\sigma(i)}) + \mathcal{L}_{reg}(\hat{p}_{\sigma(i)}, p_{\sigma(i)}).
\end{equation}

\paragraph{Our MapEX framework}

Classic frameworks do not take existing maps as input, which necessitates
introducing new modules at two key levels: at the query level we create
non-learnable \textbf{EX queries} that complement classic learnable queries (details in \cref{sec:query}), and at the matching
level we \textbf{pre-attribute} predictions to ground truths
(details in \cref{sec:matching}).

The complete MapEX framework - shown on \cref{fig:overview} -
creates non-learnable EX queries that \textbf{encode} the existing map information.
We then complete this set of EX queries with classic learnable queries to reach
a set number of queries $N\times L$. This completed set of queries is then
passed to a transformer decoder and translated into predictions by linear layers
(as usual). At \textbf{training} time, our \textbf{attribution module}
pre-attributes predictions to known ground truth correspondences before
matching, and the rest is matched normally using Hungarian Matching. The
\textbf{same loss} $\mathcal{L}$ (as in
the classic \textit{base framework}) optimizes the \textbf{same overall model}  (we add no learnable parameters). At
\textbf{test} time, the decoded non-background queries yield a HDMap
representation.

\subsection{Translating Maps into EX Queries}
\label{sec:query}

There is no mechanism in current online HDMap estimation
frameworks to account for existing map information. We therefore need
to design a new scheme that can translate existing maps into a form
understandable by standard query-based online HDMap estimation frameworks. We
propose with MapEX a simple method of encoding existing map elements into
EX queries for the decoder as shown on \cref{fig:overview}.

For a given map element, we extract $L$ evenly spaced points, with $L$ being the
number of points we seek to predict for any map element. For each point, we
craft an EX query that encodes, in the first 2 dimensions, its \textbf{map coordinates
(x,y)}, and in the next 3 dimensions a one-hot encoding of the \textbf{map element
class} (divider, crossing or boundary). The rest of the EX query is padded with
0s to reach the standard query size used by the decoder architecture.

While this query design is very simple, it presents the key benefits of both directly encoding the information of interest (point coordinates and element
class), and minimizing collisions with learned queries (thanks to the abundant
0-padding). A detailed discussion is provided in \cref{sec:ablation} with experimental comparisons to other possible designs.

Once we have $N_{EX}$ sets of $L$ queries (for the $N_{EX}$ map elements in the
existing map), we retrieve $(N-N_{EX})$ sets of $L$ assorted learnable queries
from our pool of classic learnable queries. The resulting $N\times L$ queries
are then fed to the decoder following a base classic method (e.g.
VectorMapNet, MapTRv2, ...). After we predict map elements from the queries, we can
either directly use them (at test time) or match them to the ground truth for
training.

\subsection{Map Element Pre-attribution for Training}
\label{sec:matching}

While EX queries introduce a way to account for existing map information,
nothing ensures these queries will be properly used by the model to estimate the
corresponding elements. In fact, experiments in \cref{sec:ablation} show
the network can fail to identify even fully accurate EX queries, if left on its
own. We thus introduce a pre-attribution of predictions to corresponding true map
elements before the traditional Hungarian matching used at training as shown on
\cref{fig:overview}.

Put plainly, \textbf{we keep track for each map element in the existing map of which
true map element they correspond to}: if a map element is unmodified,
shifted or warped we can tie it to the original map element in the true map. To
ensure the model learns to solely use useful information, we only keep matches
when the average point-wise displacement, between the modified map element
$m^{EX}=\{(x^{EX}_0, y^{EX}_0), \dots, (x^{EX}_{L-1}, y^{EX}_{L-1})\}$ and true
map element $m^{GT}=\{(x^{GT}_0, y^{GT}_0), \dots, (x^{GT}_{L-1},
y^{GT}_{L-1})\}$:
\begin{equation}
  s(m^{EX}, m^{GT})= \left\Vert\frac{1}{L}\sum_{i=0}^{L-1} \begin{pmatrix}x^{EX}_i \\ y^{EX}_i \end{pmatrix} - \begin{pmatrix}x^{GT}_i \\ y^{GT}_i \end{pmatrix} \right\Vert_2
\end{equation}
is below 1 meter long. In case of deletions or
additions, there are no corresponding map elements.

Given the correspondence between ground truth and predicted map elements, \textbf{we can then remove the pre-attributed map elements} from the pool of elements to be matched.
The remaining map elements (predicted and ground truth) are then matched using
some variant of the Hungarian algorithm as per usual \cite{kuhn1955hungarian,
crouse2016on}. As such, the Hungarian matching step is only needed to identify
which EX queries correspond to non-existent added map elements, and to find
classic learned queries that fit some of the true map elements absent from the
existing map (due to deletion or a strong perturbation).

Reducing how many elements must be processed by a Hungarian algorithm is
important as even the most efficient variants are of cubical complexity
$\mathcal{O}(N^3)$ \cite{crouse2016on}. This is not a major weak point in
online HDMap estimation currently as the predicted maps are small
\cite{gao2023complementing, liao2023maptrv2} ($30m\times 60m$) and only three
types of map elements are predicted. As online map generation progresses further
however, it will become necessary to accommodate an ever increasing number of
map elements as predicted maps grow both larger \cite{gao2023complementing} and
more complete \cite{li2023topology}.

\section{Experimental Results}
\label{sec:exp}

We now verify experimentally that existing maps are useful for online
HDMap estimation. After providing a general comparison of MapEX results in
relation to the literature in \cref{sec:sota}, we highlight the
improvements from using existing map information over the baseline in our
different scenarios (see \cref{sec:improvement}). We then provide deeper
understanding of the MapEX framework through careful ablations in
\cref{sec:ablation}.\\

\noindent\textit{{Setting}} We evaluate our MapEX framework on the \textbf{nuScenes dataset}
\cite{nuscenes2019} as it is the standard evaluation dataset for online
HDMap estimation. We base ourselves on the MapTRv2 framework and official
codebase. Following usual practices, we report the \textbf{Average
Precision} for each of the three map element types (divider, boundary, crossing)
at different retrieval thresholds (Chamfer distance of 0.5m, 1.0m and 1.5m)
along with overall \textbf{mean Average Precision} over the three classes. As
these averaged metrics can be difficult to interpret, we provide more granular results in
Appendix \cref{app:detailed}. To be comparable to results in the literature
\cite{liu2022vectormapnet,Ding_2023_ICCV, liao2023maptrv2}, we show results on
the nuScenes val set but conduct no hyper-parameter tuning on
the val set to avoid overfitting to it. We directly get training parameters from
MapTRv2 without tuning (using standard learning rate scaling heuristics
\cite{granziol2022learning} to adapt to our 2 GPU infrastructure). Our code,
along with the standalone MapModEX package will be made available upon
publication. A complete description of the setting is provided in Appendix
\cref{app:setting}.

For each experiment, we conduct \textbf{3 experimental runs} using three fixed random
seeds. Importantly, for a given seed and map scenario combination, the existing
map data provided during validation is fixed to facilitate comparisons. We
report results as $mean\pm std$, up to a decimal point even if standard
deviation exceeds that precision, in order to keep notations uniform.

\subsection{MapEX vs. Other Online HDMap Estimation Methods}
\label{sec:sota}

We provide in \cref{tab:sota} an overview of results from the literature
along with MapEX performance in the 5 existing map scenarios outlined in
\cref{sec:scenarios}: maps with no dividers or pedestrian crossings
(\textcolor{PineGreen}{\textbf{S1}}), noisy maps (\textcolor{Mulberry}{\textbf{S2a}} for shifted map elements, \textcolor{Mahogany}{\textbf{S2b}}
for strong pointwise noise), and substantially changed maps (\textcolor{RoyalBlue}{\textbf{S3a}} with
only those maps, \textcolor{Blue}{\textbf{S3b}} with true maps mixed in). We contextualize MapEX's
performance by comparing it both to an exhaustive inventory of existing online
HDMap estimation on comparable settings (Camera inputs, CNN Backbone) and to the
current state-of-the-art (which uses significantly more resources) for the
standard map estimation problem. While this leaves out some work
\cite{wu2023pix2map, luo2023augmenting}
on non-standard formulations, this should help contextualize our results.

\begin{table}[t]
  \centering
  \caption{\textbf{Comparison of MapEX to current methods.} In all possible
    scenarios, MapEX improves upon the base MapTRv2 model. In Scenarios \textcolor{Mulberry}{\textbf{2a}}, \textcolor{RoyalBlue}{\textbf{3a}}
    and \textcolor{Blue}{\textbf{3b}} it even beats the state-of-the-art obtained with a much stronger
    pretrained foundation backbone. Best results are highlighted in \textbf{bold}, second best are
    \underline{underlined}, third best are in \textit{italic}. ($^*$: Concurrent work,
    $^{\dagger}$: Same codebase and setting as our experiments.)}
  \label{tab:sota}
  \scalebox{0.7}{
  \begin{tabular}{lccccccc}
    \toprule
    \multirow{2}{*}{Method} & \multirow{2}{*}{Backbone} & \multirow{2}{*}{Epoch} & \multirow{2}{*}{Extra info} & \multicolumn{4}{c}{Average Precision at \{0.5m, 1.0m, 1.5m\}}  \\ \cmidrule(lr){5-8}
                            & & & & $AP_{divider}$ & $AP_{ped}$ & $AP_{boundary}$ & $mAP$ \\ \midrule
    \multicolumn{8}{c}{Previous methods} \\ \midrule
    HDMapNet \cite{li2022hdmapnet} & \cellcolor{FadedGray} EB0 & \cellcolor{LightGray} 30 & \xmark & 27.7 & 10.3 & 45.2 & 27.7 \\
    + P-MapNet$^*$ \cite{anonymous2023pmapnet} & \cellcolor{FadedGray} EB0 & \cellcolor{LightGray} 30 & Geoloc. SDMaps & 32.1 & 11.3 & 48.7 & 30.7 \\
    VectorMapNet \cite{liu2022vectormapnet} & \cellcolor{LightGray} R50 & \cellcolor{HeavyGray} 110 & \xmark & 47.3 & 36.1 & 39.3 & 40.9 \\
    + Neural Map \cite{xiong2023neural} & \cellcolor{LightGray} R50 & \cellcolor{HeavyGray} 110 & Learned map feats & 49.6 & 42.9 & 41.6 & 44.8 \\
    MapTR \cite{liao2023maptr} & \cellcolor{LightGray} R50 & \cellcolor{FadedGray} 24 & \xmark & 51.5 & 46.3 & 53.1 & 50.3 \\
    + MapVR \cite{zhang2023online} & \cellcolor{LightGray} R50 & \cellcolor{FadedGray} 24 & \xmark & 54.4 & 47.7 & 51.4 & 51.2 \\
    + Satellite Map$$ \cite{gao2023complementing} & \cellcolor{LightGray} R50 & \cellcolor{FadedGray} 24 & Geoloc. Satellite views & 55.3 & 47.2 & 55.3 & 52.6 \\
    + ADMap \cite{hu2024admap} & \cellcolor{LightGray} R50 & \cellcolor{FadedGray} 24 & \xmark & 56.2 & 49.4 & 57.9 & 54.5 \\
    PivotNet \cite{Ding_2023_ICCV} & \cellcolor{LightGray} R50 & \cellcolor{FadedGray} 24 & \xmark & 56.2 & 56.5 & 60.1 & 57.6 \\
    BeMapNet \cite{qiao2023end} & \cellcolor{LightGray} R50 & \cellcolor{LightGray} 30 & \xmark & 62.3 & 57.7 & 59.4 & 59.8 \\
    MapTRv2$^{\dagger}$ \cite{liao2023maptrv2} & \cellcolor{LightGray} R50 & \cellcolor{FadedGray} 24 & \xmark & 62.4 & 59.8 & 62.4 & 61.5 \\
    + GeMap$^{*}$ \cite{zhang2023online} & \cellcolor{LightGray} R50 & \cellcolor{FadedGray} 24 & Segmentation loss & 69.8 & 67.1 & 71.4 & 69.4 \\
    SQD-MapNet$^*$ \cite{wang2024stream} & \cellcolor{LightGray} R50 & \cellcolor{FadedGray} 24 & Prev. frames info & 66.6 & 63.6 & 64.8 & 65.0 \\
    MapNeXT$^{*}$ \cite{li2024mapnext} & \cellcolor{LightGray} R50 & \cellcolor{FadedGray} 24 & \xmark & 58.8 & 50.3 & 58.7 & 56.0 \\
    MapTRv2 \cite{liao2023maptrv2} & \cellcolor{MediumGray} V2-99 & \cellcolor{HeavyGray} 110 & Depth pretrain & 73.7 & 71.4 & 75.0 & 73.4 \\
    MapNeXT$^{*}$ \cite{li2024mapnext} & \cellcolor{HeavyGray} II-H & \cellcolor{HeavyGray} 110 & Foundation backbone & 79.3 & $\mathit{77.4}$ & 78.8 & 78.5 \\ \midrule
    \multicolumn{8}{c}{Our method} \\ \midrule
    MapEX-\textcolor{PineGreen}{\textbf{S1}} & \cellcolor{LightGray} R50 & \cellcolor{FadedGray} 24 & Map w/ only boundaries & $66.1\pm 0.6$ & $62.5\pm 0.4$ & $\mathbf{99.9\pm 0.1}$ & $76.2\pm 0.1$ \\
    MapEX-\textcolor{Mulberry}{\textbf{S2a}} & \cellcolor{LightGray} R50 & \cellcolor{FadedGray} 24 & Map w/ element shift & $\mathit{82.5\pm 1.0}$ & $\underline{78.4\pm 0.8}$ & $93.5\pm 0.4$ & $\mathit{84.8\pm 0.3}$  \\
    MapEX-\textcolor{Mahogany}{\textbf{S2b}}& \cellcolor{LightGray} R50 & \cellcolor{FadedGray} 24 & Map w/ point noise & $78.4\pm 0.1$ & $62.1\pm 0.6$ & $72.4\pm 0.4$ & $70.9\pm 0.3$  \\
    MapEX-\textcolor{RoyalBlue}{\textbf{S3a}}& \cellcolor{LightGray} R50 & \cellcolor{FadedGray} 24 & Outdated maps & $\underline{84.6\pm 0.3}$ & $74.1\pm 0.6$ & $\underline{99.1\pm 0.1}$ & $\underline{85.9\pm 0.2}$ \\
    MapEX-\textcolor{Blue}{\textbf{S3b}} & \cellcolor{LightGray} R50 & \cellcolor{FadedGray} 24 & 50\% outdated maps & $\mathbf{92.8\pm 0.1}$ & $\mathbf{87.2\pm 0.1}$ & $\underline{99.3\pm 0.2}$ & $\mathbf{93.1\pm 0.1}$ \\
    \bottomrule
  \end{tabular}
  }
\end{table}

First, it is clear from \cref{tab:sota} that \textbf{any sort of
existing map information leads MapEX to significantly outperform the
literature} on comparable settings regardless of the considered
scenario. In three out of five scenarios, existing map information even allows MapEX to
perform much better than the current state-of-the-art MapNeXt
\cite{li2024mapnext} that relies on a powerful foundation model image backbone \cite{wang2023internimage}.
Even the fairly conservative \textcolor{Mulberry}{\textbf{S2a}} scenario with imprecise map element localizations leads
to an improvement of 6.3 mAP score (i.e. 8\%).

\textbf{In all scenarios, we observe consistent improvements} over the base MapTRv2 model
in all 4 metrics. Understandably, \textcolor{Blue}{\textbf{Scenario 3b}} (with accurate existing maps half
of the time) yields the best overall performance by a large margin, thereby
demonstrating a strong ability to recognize and leverage fully accurate existing
  maps. Both \textcolor{Mulberry}{\textbf{Scenarios 2a}} (with shifted map elements) and \textcolor{RoyalBlue}{\textbf{3a}} (with ``outdated''
map elements) offer very strong overall performance with good performance for
all three types of map elements. \textcolor{PineGreen}{\textbf{Scenario 1}}, where only road boundaries are
available, shows large mAP gains thanks to its (expected) very strong retrieval
of boundaries. Even the incredibly challenging \textcolor{Mahogany}{\textbf{Scenario 2b}}, where Gaussian noise
of standard deviation 5 meters is applied to each map element point, leads to substantial gains on the base model with particularly good retrieval performance
for dividers and boundaries.

\begin{table}[t]
  \centering
  \caption{\textbf{Improvements from additional information.} In all considered
    scenarios, existing map information substantially improves results compared to
    other sources of information. ($^*$: Concurrent work, $^+$: MAE
    \cite{he2022masked} pretraining, Camera+LiDAR inputs)}
  \label{tab:improvement}
  \scalebox{0.7}{
  \begin{tabular}{lcccc}
    \toprule
    \multirow{2}{*}{Method} & \multicolumn{4}{c}{Improvement $\Delta AP = AP^{Base+Info}-AP^{Base}$}  \\ \cmidrule(lr){2-5}
                            & $\Delta AP_{divider}$ & $\Delta AP_{ped}$ & $\Delta AP_{bound}$ & $\Delta mAP$ \\ \midrule
    \multicolumn{5}{c}{Previous methods} \\ \midrule
    Neural Map & +02.3 & +06.8 & +02.6 & +03.9 \\
    Satellite Map$^*$ & +03.8 & +00.9 & +02.2 & +02.3 \\
    P-MapNet$^*$ & +04.4 & +01.0 & +03.5 & +03.0 \\
    P-MapNet$^{*,+}$ & +08.4 & +11.1 & +06.8 & +08.8 \\ \midrule
    \multicolumn{5}{c}{Our method} \\ \midrule
    MapEX-\textcolor{PineGreen}{\textbf{S1}}-onlybounds & $+03.7$ & $+02.8$ & $+37.5$ & $+14.7$ \\
    MapEX-\textcolor{Mulberry}{\textbf{S2a}}-shift-noise & $+20.1$ & $+18.6$ & $+31.1$ & $+23.3$   \\
    MapEX-\textcolor{Mahogany}{\textbf{S2b}}-point-noise & $+16.0$ & $+02.3$ & $+10.0$ & $+09.4$ \\
    MapEX-\textcolor{RoyalBlue}{\textbf{S3a}}-fullchange & $+22.2$ & $+14.3$ & $+36.7$ & $+21.4$ \\
    MapEX-\textcolor{Blue}{\textbf{S3b}}-halfchange & $+30.4$ & $+27.4$ & $+36.9$ & $+31.6$ \\
    \bottomrule
  \end{tabular}
  }
\end{table}

\subsection{Improvements Brought by MapEX over the Base Model}
\label{sec:improvement}

We now focus more specifically on the improvements that existing map information
brings to our base MapTRv2 model. For reference, we
compare MapEX gains with those brought by other sources of additional
information: \textbf{Neural Map Prior} with a global learned feature map
\cite{xiong2023neural}, \textbf{Satellite Maps} with geolocalized Satellite
views \cite{gao2023complementing}, and \textbf{P-MapNet} which uses geolocalized
SDMaps \cite{anonymous2023pmapnet}. Importantly, MapModEX relies on a stronger
base model than these methods. While this makes it harder to improve upon the
base model, it also makes it easier to reach high scores. To avoid having an
unfair advantage, we provide in \cref{tab:improvement} the absolute $\Delta
AP = AP^{Base+Info}-AP^{Base}$ score gain.

We see from \cref{tab:improvement} that using \textbf{any kind of existing map with MapEX leads to overall mAP gains larger than using any other source} of additional information (including a more sophisticated
P-MapNet setting). We generally observe very strong improvements to the model's detection performance on both lane dividers and road
boundaries. A slight exception is \textcolor{PineGreen}{\textbf{Scenario 1}}
(where we only have access to road boundaries) where the model successfully retains map information on boundaries but only provides improvements comparable
to previous methods on the two map elements it has no prior information on.
Pedestrian crossings seem to require more precise information from existing maps
as both \textcolor{PineGreen}{\textbf{Scenario 1}} and \textcolor{Mahogany}{\textbf{Scenario 2b}} (where a very destructive noise is applied to
each map point) only provide improvements comparable to existing techniques.
\textcolor{Mulberry}{\textbf{Scenarios 2a}} (with shifted elements) and \textcolor{RoyalBlue}{\textbf{3a}} (with ``outdated'' maps) lead to
strong detection scores for pedestrian crossings, which might be because these
two scenarios contain more precise information on pedestrian crossings.

\subsection{Ablations on the MapEX Framework}
\label{sec:ablation}

\begin{table}[t]
  \caption{\textbf{Influence of MapEX inputs} (sensors, maps, map element
    correspondences) on mAP. All MapEX inputs appear crucial to consistent
    overall precision.}
  \label{tab:ab_inputs}
  \centering
    \scalebox{0.675}{
      \begin{tabular}{lccccc}
        \toprule
        \multirow{2}{*}{Method} & \multicolumn{5}{c}{Mean Average Precision}  \\ \cmidrule(lr){2-6}
                                & \textcolor{PineGreen}{\textbf{S1}} & \textcolor{Mulberry}{\textbf{S2a}}-shift & \textcolor{Mahogany}{\textbf{S2b}}-noise & \textcolor{RoyalBlue}{\textbf{S3a}}-full & \textcolor{Blue}{\textbf{S3b}}-half \\ \midrule
        MapEX & $\mathbf{76.2\pm 0.1}$ & $\mathbf{84.8\pm 0.3}$ & $\mathbf{70.9\pm 0.3}$ & $\mathbf{85.9\pm 0.2}$ & $\mathbf{93.1\pm 0.1}$ \\
        ... w/o Attribution & $64.5\pm 1.9$ & $\mathbf{84.7\pm 0.7}$ & $\mathbf{72.0\pm 1.9}$ & $80.3\pm 9.6$ & $\mathbf{93.1\pm 0.2}$ \\
        ... Existing map only & $54.7\pm 0.9$ & $69.5\pm 0.5$ & $43.5\pm 0.4$ & $73.3\pm 0.3$ & $85.9\pm 0.3$ \\
        ... Sensors only (Base) & $61.4$ & $61.4$ & $61.4$ & $61.4$ & $61.4$ \\
        \bottomrule
      \end{tabular}
    }
\end{table}

\paragraph{Contribution of inputs in MapEX}

\cref{tab:ab_inputs} shows how the different types of inputs (existing maps,
map element correspondences, and sensor inputs) impact MapEX. As discussed in
\cref{sec:sota}, results demonstrate clearly that existing maps strongly
improve performance.

Line 3 of \cref{tab:ab_inputs} sets a crucial baseline by showing
the performance of \textbf{directly using the existing map} (with some corrections from a
learned model). In all scenarios, results show that using sensor inputs with MapEX substantially improves upon this baseline.


Ground truth correspondences (for pre-attribution of predictions and ground truths) seem to lower the variance of MapEX as indicated by lines 1 and 2 of \cref{tab:ab_inputs}. This demonstrates that \textbf{pre-attribution is indeed necessary to properly leverage existing map information}. A good way to understand this is
to consider our \textcolor{PineGreen}{\textbf{Scenario 1}}. In this scenario, we have access to the exact
boundary elements. With pre-attribution this consistently leads to near perfect
retrieval of those elements (see \cref{tab:sota}). This is not the case
without pre-attribution unfortunately: in two out of three runs, the network
only reaches a score below 80\% AP. This suggests pre-attribution helps ensure
MapEX consistently learns to utilize the information provided by existing maps.

\paragraph{On EX query encoding}

We use a simple encoding to translate
existing map elements into EX queries. This might be
surprising, as one might expect learned EX queries - in line with
concurrent work on map encoding \cite{luo2023augmenting, wang2024stream} - to be more useful (e.g. by projecting a 5-dimensional
vector description of the element into a query). \cref{tab:ab_query} shows
learned EX queries perform much worse than ours. Interestingly,
initializing learnable EX query with the non-learnable values might bring very
minor improvements that do not justify the added complexity.

\begin{table}[t]
  \caption{Fine grained ablation on MapEX modules (EX queries and Attribution).}
  \label{tab:ablation}
  \centering

  \begin{subtable}{0.485\linewidth}
    \centering
    \caption{\textbf{Influence of map queries} (\textcolor{PineGreen}{\textbf{Scenario 1}}). Our
      non-learnable EX query matches learnable EX
      queries while requiring no extra training.}
    \label{tab:ab_query}
    \scalebox{0.575}{
      \begin{tabular}{lcccc}
        \toprule
        \multirow{2}{*}{Method} & \multicolumn{4}{c}{Average Precision at \{0.5m, 1.0m, 1.5m\}}  \\ \cmidrule(lr){2-5}
                                & $AP_{divider}$ & $AP_{ped}$ & $AP_{boundary}$ & $mAP$ \\ \midrule
        MapEX encoding & $66.1\pm 0.6$ & $62.5\pm 0.4$ & $99.9\pm 0.1$ & $76.2\pm 0.1$ \\
        Linear encoding & $63.4\pm 0.3$ & $61.1\pm 0.3$ & $100 \pm 0.1$ & $74.8\pm 0.1$ \\
        Lin. enc. w/ MapEx init. & $66.6\pm 0.1$ & $62.5\pm 0.9$ & $100\pm 0.1$ & $76.4\pm 0.3$ \\
        \bottomrule
      \end{tabular}
    }
  \end{subtable}
  \hfill
  \begin{subtable}{0.485\linewidth}
    \centering
    \caption{\textbf{Influence of pre-attribution} (\textcolor{Mulberry}{\textbf{Scenario 2a}}).
      Selecting which existing map elements to pre-attribute
      significantly improves results.}
    \label{tab:ab_attribution}
    \scalebox{0.675}{
      \begin{tabular}{lcccc}
        \toprule
        \multirow{2}{*}{Method} & \multicolumn{4}{c}{Average Precision at \{0.5m, 1.0m, 1.5m\}}  \\ \cmidrule(lr){2-5}
                                & $AP_{divider}$ & $AP_{ped}$ & $AP_{boundary}$ & $mAP$ \\ \midrule
        MapEX & $82.5\pm 1.0$ & $78.4\pm 0.8$ & $93.5\pm 0.4$ & $84.8\pm 0.3$ \\
        ... w/o sim. thresh. & $79.5\pm 1.6$ & $76.4\pm 0.9$ & $91.9\pm 0.2$ & $82.6\pm 0.7$ \\
      \bottomrule
    \end{tabular}
}
  \end{subtable}
\end{table}

\paragraph{On ground truth pre-attribution}

Since pre-attributing map elements is important to consistently use existing map information (see \cref{tab:ab_inputs}), it might be
tempting to pre-attribute all the corresponding map elements instead of filtering them like we do in MapEX. \cref{tab:ab_attribution} shows that discarding correspondences when the existing map element is too different (see \cref{sec:matching}) does lead to stronger performance than
indiscriminate attribution. In essence, it is preferable
to use a learnable query instead of EX queries when the existing map element is too
different from the ground truth.

\section{Discussion}

We propose to improve online HDMap estimation by taking advantage of an overlooked
resource: \textbf{existing maps}. To study this, we outline three \textbf{realistic scenarios}
where existing (minimalist, noisy or outdated) maps are available and introduce
a new \textbf{MapEX} framework to leverage these maps. As current frameworks cannot take existing maps as inputs, we
develop two novel modules: one encoding map elements into \textbf{EX queries}, and
another that \textbf{pre-attributes} predictions to known ground truth correspondences to ensure the model leverages these queries.

Experimental results demonstrate that existing maps represent a crucial information for online HDMap estimation, with MapEX significantly improving upon
comparable methods regardless of the scenario. In fact, the median scenario (in
terms of mAP) - \textcolor{Mulberry}{\textbf{Scenario 2a}} with randomly shifted map elements - improves upon
the base MapTRv2 model by \textbf{38\%} and upon the current state-of-the-art by \textbf{8\%}.

We hope this work will lead new online HDMap estimations to account for
existing information. Existing maps - good or bad - are \textbf{widely available}. To
ignore them is to forego a \textbf{crucial} tool in reliable online HDMap
estimation.

\paragraph{Limitations}
By nature, MapEX makes use of an underlying map estimation method to function without
providing one by itself. Although the formalism should be compatible with most
modern methods, the exact mechanisms introduced could
require modifications to accommodate new query schemes (e.g. for new map elements). Our main claim that
using existing maps largely improves map estimation should remain relevant in
any case however.

\paragraph{Acknowledgements} This work was realized by the MultiTrans project
funded by the Agence Nationale de la Recherche under grant reference
ANR-21-CE23-0032. The authors are grateful to the OPAL infrastructure from
Université Côte d'Azur for providing resources and support.

\bibliographystyle{splncs04}
\bibliography{main}

\clearpage

\setcounter{page}{1}
\section*{Supplementary material}

\begin{figure}[H]
  \centering
  \includegraphics[width=0.9\linewidth]{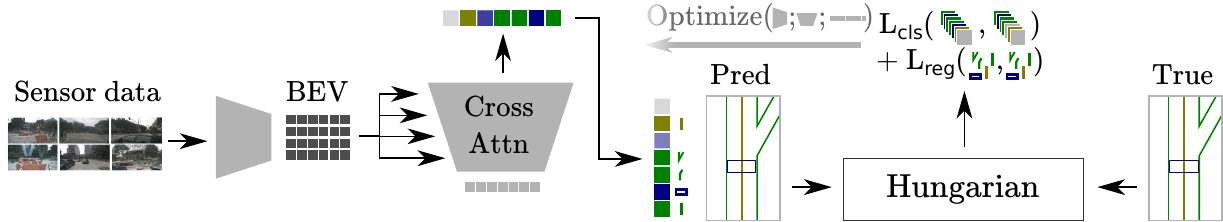}
  \caption{\textbf{Overview of a classic query based framework}. Sensor data is
    encoded into BEV features, before being cross attended with learned
    detection queries in a DETR-like scheme. The final attended queries serve to
  predict coordinates and classes of map elements. The model is trained using a
  Hungarian matching between predictions and ground truths.}
  \label{app:classic}
\end{figure}

We provide in this Appendix some additional details to understand our work:
\begin{itemize}
\item We provide more details on our experimental setting in Sec.~\ref{app:setting}.
\item We discuss how the Argoverse 2 Trust but Verify relates to our problem in Sec.~\ref{app:tbv}.
\item We provide the detailed precision tables and qualitative examples for our main results in Sec.~\ref{app:detailed}.
\item We study how the model behaves with exact map inputs in Sec.~\ref{app:change}.
\item We give pseudocode overviews of our two original MapEX modules in Sec.~\ref{app:pseudocode}.
\item We give a figure of a query based onlline HDMap estimation framework
  without MapEX modules in Fig.~\ref{app:classic}.
\end{itemize}

\section{Detailed setting and codebase}
\label{app:setting}

We introduce here the detailed experimental details used for our experiments
along with in-depth explaination of how existing maps are obtained for our
various scenarios. Our code is largely based on the official MapTRv2
code\footnote{\url{https://github.com/hustvl/MapTR/tree/maptrv2}}, and will be
made available along with our standalone MapModEX libray upon acceptance of this paper.

\paragraph{Training details}

We largely reprise the 24 epochs training settings from our MapTRv2
\cite{liao2023maptrv2} base, which were described in the original paper as:

\begin{quotation}
``ResNet50 is used as the image back-
bone network unless otherwise specified. The optimizer
is AdamW with weight decay 0.01. The batch size is 32
(containing 6 view images) and all models are trained with 8
NVIDIA GeForce RTX 3090 GPUs. Default training schedule
is 24 epochs and the initial learning rate is set to 6 × 10-4
with cosine decay. We extract ground-truth map elements in
the perception range of ego-vehicle following [...]
The resolution of source nuScenes images is 1600 × 900.
[...] Color jitter is used by default in both nuScenes
dataset and Argoverse2 dataset. The default number of
instance queries, point queries and decoder layers is 50,
20 and 6, respectively. For PV-to-BEV transformation, we
set the size of each BEV grid to 0.3m and utilize efficient
BEVPoolv2 [77] operation. Following [16], $\lambda_c$ = 2, $\lambda_p$ = 5,
$\lambda_d$ = 0.005. For dense prediction loss, we set $\alpha_d, \alpha_p, \alpha_b$
to 3, 2 and 1 respectively. For the overall loss, $\beta_o$ = 1,
$\beta_m$ = 1, $\beta_d$ = 1.''
\end{quotation}

Our own training setting solely differs from MapTRv2's in the fact that we train
on 2 NVIDIA Quadro RTX 8000 GPUs. This in turn mean we need to reduce the batch
size by 4 and scale learning rates by 2 following standard scaling heuristics
for Adam optimizers \cite{granziol2022learning}.

\paragraph{\textcolor{PineGreen}{\textbf{Scenario 1}} implementation} We remove the
divider and pedestrian crossings from available HDMaps.

\paragraph{\textcolor{Mulberry}{\textbf{Scenario 2a}} implementation} For each map element localization, we add
noise from a Gaussian distribution with standard deviation of 1 meter. This has
the effect of applying a uniform translation to each map element (dividers,
boundaries, crosswalks).

\paragraph{\textcolor{Mahogany}{\textbf{Scenario 2b}} implementation}
For each ground truth point - keeping in mind a map
element is made up of 20 such points - we sample noise from a Gaussian
distribution with standard deviation of 5 meters and add it to the point
coordinates.

\paragraph{\textcolor{RoyalBlue}{\textbf{Scenario 3a}} implementation}
We delete 50\% of the pedestrian crossings and lane dividers in the map, add a
few pedestrian crossings (half the amount of the remaining crossings) and
finally apply a small warping distortion to the map. The warping distortion is
composed of first trigonometric warping with horizontal and vertical amplitudes
1, and inclination 3. We then perform triangular warping following a slightly
perturbed grid where each point on the regular grid is shifted according to
random Gaussian noise with standard deviation 1.

\paragraph{\textcolor{Blue}{\textbf{Scenario 3b}} implementation}
For each map, we draw a uniform random value between 0 and 1. If it is below p=0.5
we keep the true HDMap, otherwise we perturb it in the same way as in Scenario 3a.

\section{On the Trust but Verify dataset}
\label{app:tbv}

The Argoverse 2 Trust but Verify (TbV) dataset \cite{Lambert21neurips} offers
situations where the HDMap does not fit sensor inputs for change detection.
Unfortunately, it is \textbf{not suitable} for our purposes it only says whether
the current map fits sensor data (yes or no) \textbf{without
giving the new map} (see Sec.~3.3 of \cite{Lambert21neurips} or the associated
code). Without the relevant ground truth we cannot evaluate on it.

Additionally, while TbV is an excellent dataset for change detection, it
unfortunately contains a limited number of real scenarios to train model for
online HDMap acquisition. Moreover, a number of the change scenarios are
indiscernible for our HDMap representation (e.g. change in the type of divider).
Interestingly, the limited number of hand curated change situations is reserved
for the validation and test sets with the train set generated from synthetic
data. Where TbV chooses to generate synthetic views that differ from the
available HDMap, we take the opposite view of modifying the HDMaps. While this
is likely less desirable for change detection, it is of no consequence for
online HDMap acquisition and much lighter computationally.

\section{Fine grained results of map estimations}
\label{app:detailed}

\cref{tab:detailed} provides a deeper look into the detailed results of MapEX and sheds light on how the different types of existing maps actually benefit the model.

Interestingly, the noisy \textcolor{Mulberry}{\textbf{Scenarios 2a}} and \textcolor{Mahogany}{\textbf{2b}} seem to help the model give a rough approximation of map elements (good scores for retrieval thresholds of 1.5m) but are less useful when it comes to predict very precise element localizations. As such, these scenarios appear to help the model by providing a \textbf{general idea of what the situation looks like}. Nevertheless, \textcolor{Mulberry}{\textbf{Scenarios 2a}} appears to still subtantially improve the base MapTRv2 model for precise element localizations at 0.5m (which is much lower than the standard deviation of the added noise).

Conversely, when outdated map \textcolor{RoyalBlue}{\textbf{Scenarios 3a}} and \textcolor{Blue}{\textbf{3b}} are useful to predict map elements, they tend to provide fairly precise element localizations (the gap between precision at 0.5m and 1.5m is lower). While these scenarios strongly improve performance at all precision thresholds, the improvement is also much larger for very precise element localizations. As such, they seem to work by providing a \textbf{more precise approximations of map elements}.

\textcolor{PineGreen}{\textbf{Scenario 1}} (with only boundaries) for its part shines by providing near perfect estimations of map boundaries at all levels: it properly \textbf{identifies the provided road boundary localizations as perfectly accurate} and restitutes them as is. Interestingly, it also provides significant gains in precision at all retrieval thresholds for lane dividers and pedestrian crossings even though the existing map has no information on them.

\begin{table}
  \centering
  \caption{Detailed table of retrieval Precisions at different thresholds for the main results. Reproduced scores for the base MapTRv2 model are given for reference.}
  \label{tab:detailed}
  \begin{subtable}{0.475\linewidth}
    \caption{}
  \scalebox{0.65}{
  \begin{tabular}{lccccccccc}
    \toprule
    \multirow{2}{*}{Method} & \multicolumn{3}{c}{$AP_{divider}$}  \\ \cmidrule(lr){2-4} 
                            & $Precision_{divider}^{0.5}$ & $Precision_{divider}^{1.0}$ & $Precision_{divider}^{1.5}$ \\ \midrule
    MapTRv2 & $46.0$ & $66.4$ & $75.4$  \\
    MapEX-\textcolor{PineGreen}{\textbf{S1}} & $50.7 \pm 0.3$ & $69.6 \pm 0.7$ & $77.8 \pm 0.6$  \\
    MapEX-\textcolor{Mulberry}{\textbf{S2a}} & $\mathit{62.8 \pm 2.1}$ & $\mathit{83.6 \pm 1.4}$ & $\underline{92.1 \pm 1.3}$  \\
    MapEX-\textcolor{Mahogany}{\textbf{S2b}} & $50.9 \pm 3.0$ & $77.5 \pm 2.2$ & $89.0 \pm 1.0$ \\
    MapEX-\textcolor{RoyalBlue}{\textbf{S3a}} & $\underline{76.2 \pm 0.5}$ & $\underline{86.6 \pm 0.3}$ & $\mathit{90.8 \pm 0.3}$ \\
    MapEX-\textcolor{Blue}{\textbf{S3b}} & $\mathbf{88.4 \pm 0.5}$ & $\mathbf{93.8 \pm 0.4}$ & $\mathbf{95.8 \pm 0.2}$ \\
    \bottomrule
  \end{tabular}
}
  \end{subtable}
  \hfill
  \begin{subtable}{0.475\linewidth}
    \caption{}
  \scalebox{0.65}{
  \begin{tabular}{lccccccccc}
    \toprule
    \multirow{2}{*}{Method} & \multicolumn{3}{c}{$AP_{ped}$} \\ \cmidrule(lr){2-4} 
                            & $Precision_{ped}^{0.5}$ & $Precision_{ped}^{1.0}$ & $Precision_{ped}^{1.5}$ &  \\ \midrule
    MapTRv2 & $34.5$ & $65.1$ & $78.8$  \\
    MapEX-\textcolor{PineGreen}{\textbf{S1}} &  $38.8 \pm 0.5$ & $68.8 \pm 0.5$ & $80.0 \pm 0.5$ \\
    MapEX-\textcolor{Mulberry}{\textbf{S2a}} &  $\mathit{46.5} \pm 0.5$ & $\underline{85.5 \pm 1.9}$ & $\mathbf{97.2 \pm 0.4}$ \\
    MapEX-\textcolor{Mahogany}{\textbf{S2b}} &  $28.4 \pm 0.5$ & $72.3 \pm 2.1$ & $\mathit{91.7 \pm 1.8}$ \\
    MapEX-\textcolor{RoyalBlue}{\textbf{S3a}} &  $\underline{56.2 \pm 0.4}$ & $\mathit{79.4 \pm 0.5}$ & $86.7 \pm 0.7$ \\
    MapEX-\textcolor{Blue}{\textbf{S3b}} &  $\mathbf{77.7 \pm 0.5}$ & $\mathbf{89.9 \pm 0.3}$ & $\underline{93.6 \pm 0.3}$ \\
    \bottomrule
  \end{tabular}
}
    \vspace{0.5cm}
  \end{subtable}

  \begin{subtable}{0.475\linewidth}
  \centering
    \caption{}
    \scalebox{0.65}{
  \begin{tabular}{lccccccccc}
    \toprule
    \multirow{2}{*}{Method} &  \multicolumn{3}{c}{$AP_{boundary}$} \\ \cmidrule(lr){2-4}
                            & $Precision_{boundary}^{0.5}$ & $Precision_{boundary}^{1.0}$ & $Precision_{boundary}^{1.5}$  \\ \midrule
    MapTRv2 & $39.6$ & $70.3$ & $80.6$  \\
    MapEX-\textcolor{PineGreen}{\textbf{S1}} & $\mathbf{99.8 \pm 0.1}$ & $\mathbf{99.8 \pm 0.1}$ & $\mathbf{100.0 \pm 0.1}$ \\
    MapEX-\textcolor{Mulberry}{\textbf{S2a}} &  $80.5 \pm 0.9$ & $96.4 \pm 0.4$ & $98.9 \pm 0.2$ \\
    MapEX-\textcolor{Mahogany}{\textbf{S2b}} &  $34.9 \pm 0.2$ & $75.8 \pm 0.2$ & $90.0 \pm 0.3$ \\
    MapEX-\textcolor{RoyalBlue}{\textbf{S3a}} &  $\underline{97.4 \pm 0.3}$ & $\mathbf{99.9 \pm 0.1}$ & $\mathbf{100.0 \pm 0.1}$\\
    MapEX-\textcolor{Blue}{\textbf{S3b}} &  $\underline{97.8 \pm 0.4}$ & $\mathbf{99.7 \pm 0.3}$ & $\mathbf{100.0 \pm 0.1}$\\
    \bottomrule
  \end{tabular}
    }
  \end{subtable}
\end{table}

\section{Map change detection}
\label{app:change}

We discuss here an additional module initially explored for MapEX. We include this discussion here as this module does not improve performance (and is therefore not an improved version of MapEX), but sheds light on what happens when perfect existing maps are available to the model.

\subsection{Map change detector}
\label{app:changifier}

There are a number of situations where fully accurate HDMaps might be mixed in
with the imperfect HDMaps (e.g. our \textcolor{Blue}{\textbf{Scenario 3b}}). As such, we propose a
lightweight change detection module to leverage these situations.

We introduce a learned change detection query token and perform cross-attention
between this token and intermediate map element queries at different stages of
the decoder. This token is then decoded by dense layer into a change prediction
$c\in [0,1]$ (with a sigmoid activation). At training time, we train this token
with a binary cross entropy loss (with target $\hat{c}=1$
if the map is not fully accurate and $\hat{c}=0$ if it is): we minimize
\begin{equation}
  \mathcal{L}=\mathcal{L}_{Base}+\mathcal{L}_{BCE}(c,\hat{c}),
\end{equation}
with $\mathcal{L}_{Base}$ the loss of the base online HDMap estimator. At test
time, if no change is detected we output the existing HDMap instead of the
prediction (and we output the decoder predictions as usual if a change is detected).

Using the existing HDMap has two benefits: it provides a very precise
HDMap (something most methods struggle with \cite{Ding_2023_ICCV}), and it
provides a way to stop the map estimation process early. Indeed, returning the
existing map removes the need for further decoding of the query tokens which can
be expensive.


\subsection{Processing accurate existing maps}
\label{sec:ab_change}

We take a closer look at how MapEX deals with perfectly accurate
existing maps as it can sometimes happen in scenarios like \textcolor{Blue}{\textbf{Scenario 3b}}. To this
end, we compare MapEX to variants that use an explicit map change detection
module (described in Appendix \ref{app:change}) and substitute the predicted map
with the input existing map if no change is detected. Tab.~\ref{tab:ab_change}
shows MapEX does not need a change detection module: it recognizes and uses
accurate existing map elements on its own. In fact, training a change detection
module jointly with MapEX appears to deteriorate performance.

\begin{table}[H]
  \centering
  \caption{Usefulness of the change detection module (\textcolor{Blue}{\textbf{Scenario 3b}}). MapEX seems
    to recognize and leverage existing maps without the need for explicit
    change detection.}
  \label{tab:ab_change}
  \scalebox{0.7}{
    \begin{tabular}{lcccc}
      \toprule
      \multirow{2}{*}{Method} & \multicolumn{4}{c}{Average Precision at \{0.5m, 1.0m, 1.5m\}}  \\ \cmidrule(lr){2-5}
                              & $AP_{divider}$ & $AP_{ped}$ & $AP_{boundary}$ & $mAP$ \\ \midrule
      MapEX & $92.8\pm 0.1$ & $87.2\pm 0.1$ & $99.3\pm 0.2$ & $93.1\pm 0.1$ \\
      ... w/ substitution & $92.5\pm 0.3$ & $87.3\pm 0.3$ & $99.4\pm 0.1$ & $93.0\pm 0.1$ \\
      ... w/ sub. \& optimization & $92.5\pm 0.2$ & $87.2\pm 0.2$ & $99.3\pm 0.1$ & $93.0\pm 0.1$ \\
      \bottomrule
    \end{tabular}
  }
\end{table}


\section{Pseudo code}
\label{app:pseudocode}

We provide here pseudo code for our two additional modules: the EX query
encoding module (Alg.~\ref{alg:query}) and the pre-attribution code (Alg.~\ref{alg:attribution}).

\begin{algorithm}
  \KwData{Map element $m^{EX}=\{(x^{EX}_0, y^{EX}_0), \dots, (x^{EX}_{L-1},
    y^{EX}_{L-1})\}$ of class $c$ (among divider, crossing and boundary).}
  \KwResult{list query\_list of $L$ $H$-dimensional EX queries.}
  query\_list = []\;
  \For{$i\gets 0$ \KwTo L-1}{
    \tcc{Encode position}
    pos\_vec = array([$x^{EX}_i, y^{EX}_i$])\;
    \tcc{Encode class}
    class\_vec = one\_hot($c$, num\_class=3)\;
    \tcc{Build query}
    pad\_vec = zeros($H-5$)\;
    query\_i = concatenate([pos\_vec, class\_vec, pad\_vec])\;
    query\_list.append(query\_i)\;
  }
  \Return query\_list\;
  \caption{Encoding map elements into EX Queries.}
  \label{alg:query}
\end{algorithm}

\begin{algorithm}
  \KwData{Predictions $p=\{p_i\}_{i=0,...,49}$, (Padded) ground truths
    $g=\{g_i\}_{i=0,...,49}$, correspondence list $c=\{c_i\}_{i=0,...,49}$ where
  $c_i=-1$ if there is no correspondence}
  \KwResult{Assignment $a=\{a_i\}_{i=0,...,49}$ where $a_i$ is the index of the
    ground truth associated to the $i$-th prediction.}
    \tcc{Split off pre-attributed pairs}
    $p^p,g^p,c^p,p^n,g^n,c^n,split\_inds$ = Split(p,g,c)\;
    \tcc{Perform Hungarian matching}
    $a^n$ = Hungarian($p^n,g^n$)\;
    \tcc{Merge $c^p$ with $a^n$}
    a = Merge($p^p$, $a^n$, split\_inds);
    \Return a\;
  \caption{Hungarian matching with pre-attribution.}
  \label{alg:attribution}
\end{algorithm}

\clearpage
\newpage

\end{document}